\documentclass{Interspeech2024}
\usepackage{multirow}
\usepackage{tabularx}
\usepackage{graphicx}
\usepackage{stfloats}
\usepackage{svg}
\usepackage{url}
\usepackage{authblk}
\usepackage{hyperref}

\interspeechcameraready

\title{Towards an End-to-End Framework for Invasive Brain Signal Decoding with Large Language Models}

\renewcommand{\thefootnote}{\fnsymbol{footnote}}
\name[affiliation={1\dagger}]{Sheng}{Feng}
\name[affiliation={1\dagger}]{Heyang}{Liu}
\name[affiliation={1,2\ast}]{Yu}{Wang}
\name[affiliation={1,2}]{Yanfeng}{Wang}

\address{
  $^1$Cooperative Medianet Innovation Center, Shanghai Jiao Tong University, China\\
  $^2$Shanghai Artificial Intelligence Laboratory, China}
\email{\{fs0015,liuheyang,yuwangsjtu,wangyanfeng622\}@sjtu.edu.cn}


\begin{document}

\maketitle

\renewcommand{\thefootnote}{\fnsymbol{footnote}}
\setcounter{footnote}{0}
\footnotetext{$^\dagger$Equal contribution}
\footnotetext{$^\ast$Corresponding author}
\footnotetext{This work is supported by National Key R\&D Program of China (No. 2022ZD0162101), National Natural Science Foundation of China (No. 62106140) and STCSM (No. 21511101100, No. 22DZ2229005).}

\begin{abstract}

    In this paper, we introduce a groundbreaking end-to-end (E2E) framework for decoding invasive brain signals, marking a significant advancement in the field of speech neuroprosthesis. Our methodology leverages the comprehensive reasoning abilities of large language models (LLMs) to facilitate direct decoding. By fully integrating LLMs, we achieve results comparable to the state-of-the-art cascade models. Our findings underscore the immense potential of E2E frameworks in speech neuroprosthesis, particularly as the technology behind brain-computer interfaces (BCIs) and the availability of relevant datasets continue to evolve. This work not only showcases the efficacy of combining LLMs with E2E decoding for enhancing speech neuroprosthesis but also sets a new direction for future research in BCI applications, underscoring the impact of LLMs in decoding complex neural signals for communication restoration. Code will be made available at  \href{https://github.com/FsFrancis15/BrainLLM}{https://github.com/FsFrancis15/BrainLLM}.
    
\end{abstract}

\section{Introduction}

Language is a unique mode of communication specific to humans \cite{mehler2006language, pagel2017q}. With the advancements in brain activity recording methods and the evolution of deep learning models, decoding textual information from brain recordings has become feasible. Previous research has demonstrated that the human brain contains a vast amount of acoustic representations and semantic information \cite{chang2010categorical, mesgarani2014phonetic}, while the relevance of pre-trained speech and language models to brain representations has been proven \cite{millet2022toward, oota2023neural}. Utilizing deep learning methods to model the paradigms of brain neural activity has become the mainstream approach. 

The goal of brain decoding is to decode the language stimuli perceived or the information being expressed from brain activity \cite{D_fossez_2023}. A more fine-grained task, called speech neuroprosthesis, directly decodes the words a subject intends to say \cite{10.3389/fnins.2015.00217,doi:10.1056/NEJMoa2027540}. The participants typically lose the speech ability due to illness or injury, but their brains possess intact language centers capable of phonation. Speech neuroprosthesis can significantly improve the communication methods of patients with speech disorders, as well as prompt neuroscientific researchers to enhance their understanding of the brain's language functions while also pioneering new directions for brain-computer interface (BCI).

Speech neuroprostheses share a common goal with Automatic Speech Recognition (ASR) technologies: both aim to accurately transcribe textual sequences from temporal signals. Recent advancements have adopted an approach inspired by hybrid ASR models, which utilize neural networks to model brain activities associated with speech acoustics \cite{willett2023high, metzger2023high}. These models integrate acoustic scores with linguistic scores derived from n-gram language models trained on vast corpora. The generation of decoding hypotheses is facilitated through the Viterbi search algorithm. This pioneering method has achieved notable decoding accuracy from brain activity, with a word error rate of 23.8\% across a substantial vocabulary size of 125,000 words. 

Despite the exciting results obtained with cascaded recognition schemes, we are more inclined towards sequential decoding with an E2E approach. This paradigm learns the mapping directly from raw brain recordings to text transcriptions, training the entire system through a unified framework, which greatly simplifies the training process. It has the potential to learn optimal feature representations from qualitative data and surpass the performance of traditional hybrid models on certain tasks. The reasoning capabilities and the ability to integrate multiple modalities demonstrated by LLMs further amplify the advantages of E2E models, a point that has been fully exemplified in ASR \cite{zhang2023speechgpt, fathullah2023prompting}. These factors have prompted a shift towards a sequential approach in speech neuroprosthesis. Based on this, we have implemented an E2E brain decoding system with LLMs. Our system utilizes a lightweight feature extractor that transforms invasive brain recordings into feature representations, coupled with an LLM-based decoder as its core. It employs a multi-stage training process preceding hypothesis decoding. The initial stage aligns modalities by mapping brain recording features to the space of LLM tokens, followed by a finetuning stage that optimizes the decoder's ability to process intermediate representations. Our contributions are as follows:

\begin{itemize}
    \item We introduce the first system capable of decoding inner speech from brain activity in an E2E fashion, offering new perspectives in speech neuroprosthesis research.
    \item By harnessing the inference capabilities of LLMs and eliminating the need for external n-gram language models, our system achieves performance on par with hybrid models.
    \item Our work includes the finetuning of various LLMs for the processing and decoding of brain activity, offering a valuable benchmark for the development of speech-functional BCIs.
\end{itemize}

\section{Related Works}

\subsection{Speech neuroprosthesis}

The goal of speech neuroprosthesis is to decode sentences that participants intend to speak directly from their brain signals. Participants in this research usually lose their ability to speak due to neurological damage, but their brains remain intact, making precise brain-to-text decoding possible. This technology represents a promising pathway for developing voice communication aids, benefiting those who lose their speaking ability due to damage to vocal organs or neurological defects.

Phonemes are the basic units of pronunciation, and the earliest work involved recognizing a small set based on brain activity \cite{zhao2015classifying, stavisky2018decoding}. Subsequently, researchers achieved brain decoding with words as units, but still within the constraints of a small vocabulary, while also needing to ensure the distinctiveness in the pronunciation of the selected words and minimite words in each trial \cite{ mohanchandra2016communication,gonzalez2017sonification}. The low signal-to-noise ratio (SNR) of brain activity recordings is one of the key factors limiting the decoding space, and the aforementioned works are based on non-invasive or semi-invasive recordings. Although these approaches are more portable and riskless, the signal quality is 20-100 times worse than that simultaneously obtained in invasive methods \cite{ball2009signal}. This represents a tendency in brain activity recording: when decoding continuous sentences with a fine granularity over a large vocabulary, invasive recordings may be necessary. In this mode, brain-to-text can be achieved with vocabularies of several dozen words, and simple sentence decoding can be achieved \cite{moses2021neuroprosthesis}. The first research to achieve large-vocabulary brain decoding draws inspiration from hybrid model speech recognition, completing phoneme recognition of brain activity through a custom network architecture and utilizing a language model trained with Kaldi \cite{povey2011kaldi} for search decoding \cite{willett2023high, metzger2023high}. Such an approach requires the support of language models trained on extensive text. The shift from cascaded models to E2E systems not only represents the development of ASR but is also likely the preference for speech neuroprosthesis.

\subsection{Automatic speech recognition}

Cascaded hybrid models and E2E models represent two paradigms in speech recognition. The former, appearing earlier, includes an acoustic model that calculates pronunciation probabilities, a language model that models the distribution of transcription, and a predefined lexicon \cite{dahl2011context, peddinti2015time}. The scores from the acoustic model and the language model together determine the decoding hypothesis. E2E models directly fit the mapping between speech waves and the corresponding transcriptions, fully utilizing the network's capability to extract deep features, gradually replacing hybrid models as the mainstream approach. This method does not require traditional preprocessing steps such as acoustic feature extraction, forced alignment, etc., but follows a sequence-to-sequence method. Applying the E2E concept to speech neuroprosthesis holds great potential.

\section{Proposed Method}

\subsection{End-to-End brain-to-text framework}




A brain signal for a given period is represented as a multivariate time series $\boldsymbol{X} \in \mathbb{R}^{T \times F}$, where $T$ is the number of timesteps and $F$ is the feature dimension. The text transcription is a sequence of words $Y \in \mathcal{V}^{L}$ of length $L$ on an open vocabulary $\mathcal{V}$, where $L \ll T$. The goal of the brain-to-text task is to discover a mapping between brain signals and corresponding text sequences $f$, such that $\hat{Y} = f(\boldsymbol{X})$.

Inspired by the recent success of the LLMs, we implement an E2E framework to handle this cross-modality recognition task. First, the brain signal is sent to a modality-specific feature extractor to obtain high-quality features. These feature embeddings are then mapped into the space of the LLM tokens, which allows us to decode the corresponding text hypothesis.

\begin{figure}[h]
    \centering
    \includegraphics[width=0.45\textwidth]{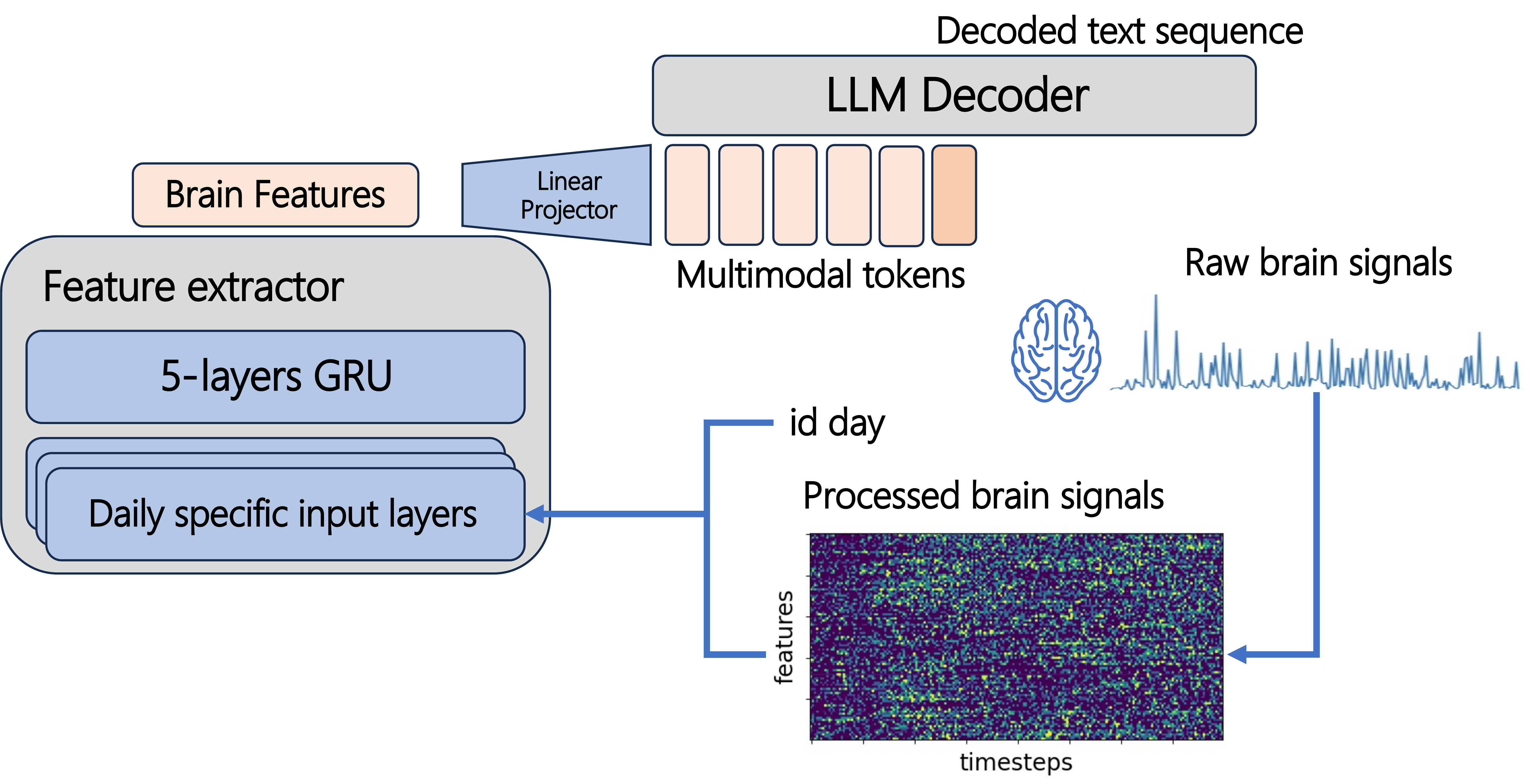}
    \caption{Diagram of the proposed end-to-end invasive brain signal decoding framework. The multimodal tokens generated by the feature extractor from processed brain signals are used for LLM decoding.}
    \label{E}
\end{figure}

The overall framework is illustrated in Figure~\ref{E}. The recognition strategy $f$ is decomposed by $f = f_{LLM} \circ g \circ f_{brain}$. Where $f_{brain}$ is brain feature extractor that maps the brain signal into an intermediate representation $\boldsymbol{Z} = f_{brain}(\boldsymbol{X}) \in \mathbb{R}^{T_b \times F_b}$ of different time length $T_b$ and feature dimension $F_b$, the function $g$ is a linear transformation that transforms the feature space into the dimension of LLM embeddings $\boldsymbol{E} = g(\boldsymbol{Z}) = \boldsymbol{ZM}+\boldsymbol{M_0}$, which can be viewed as a sequence of token embeddings: $\boldsymbol E = [\boldsymbol{e}_1,...,\boldsymbol{e}_{T_b}] \in \mathbb{R}^{T_b \times D}$. The LLM decoder takes the transformed token embeddings and autoregressively decodes word sequence: $\hat{Y} = f_{LLM}([\boldsymbol{E},\boldsymbol{e}_{bos}])$.

\subsection{Feature extractor}

In our framework, the modality-specific feature extractor should have a good functional form to capture the information of brain signals. We use a well-designed 5-layer GRU network with a 1024 hidden size, following the acoustic part of the previous hybrid method \cite{willett2023high}. Additionally, the feature extraction module integrates a specialized input network tailored for daily variations, designed to adapt to the temporal distribution changes in brain signals. This is further enhanced by a common GRU back-end, ensuring comprehensive and dynamic signal analysis.

To facilitate the training process, we first pre-train the feature extractor to map the brain signals directly into the sub-word level of the ground truth sentences before plugging it into the framework. We design two sequence-to-sequence tasks to pre-train the feature extractor. The first task is the brain signals to phonemes task, which has shown success in previous works \cite{willett2023high}. We decompose the sentence into a sequence of phonemes by using the python g2p-en package \cite{g2pE2019} before training the feature extractor to output the phoneme sequences given the brain recordings. The second is similar in the task form but replaces the phoneme with sub-word units, like byte pair encoding (BPE) \cite{gage1994new, sennrich-etal-2016-neural}. We first train the BPE model on the transcriptions in the training dataset to obtain the best sub-word decomposition using SentencePiece package\cite{kudo2018sentencepiece}. In both tasks, we add a linear classification head to the feature extractor and then train it by connectionist temporal classification (CTC) \cite{graves2006connectionist}.


\subsection{LLM decoder}

Choosing the LLM backbone is critical in a cross-modality E2E framework. The LLM serves as the decoder for the final text output, and its ability to understand different modalities has a significant impact. We study three types of well-known open-source LLMs in our framework and further investigate how they influence the overall performance.

\textbf{GPT-2}: GPT-2, developed by OpenAI \cite{radford2019language}, is a generative large language model (LLM). Its emergence marked a significant point in the research of large-scale models, setting the groundwork for the subsequent rapid development of generative large language models. In our work, we utilize the GPT-2 version with 117M parameters as one of our LLM decoders.

\textbf{OPT}: Openly Pre-trained Transformers (OPT) \cite{zhang2022opt}, developed by Meta AI, is a generative large language model. Its goal is to match the size and performance of the renowned GPT-3 model. OPT employs a decoder-only transformer architecture, with parameters varying from 175M to 175B. In our work, we opted for the 1.3B, 2.7B, and 6.7B versions of OPT as our large language model decoders.

\textbf{Llama 2}: Llama 2 \cite{touvron2023llama}, one of Meta AI's latest generative Language Models (LLMs), has parameters ranging from 7B to 70B. It outperforms most other open-source generative LLMs on various benchmarks. Notably, the Llama 2 model is frequently used in multi-modal frameworks, which demonstrates its strong ability to understand representations in different modalities. In our work, we've selected the 7B version of Llama 2 as our LLM decoder.

\subsection{Multi-stage training}

\begin{figure}[h]
    \centering
    \includegraphics[width=0.5\textwidth]{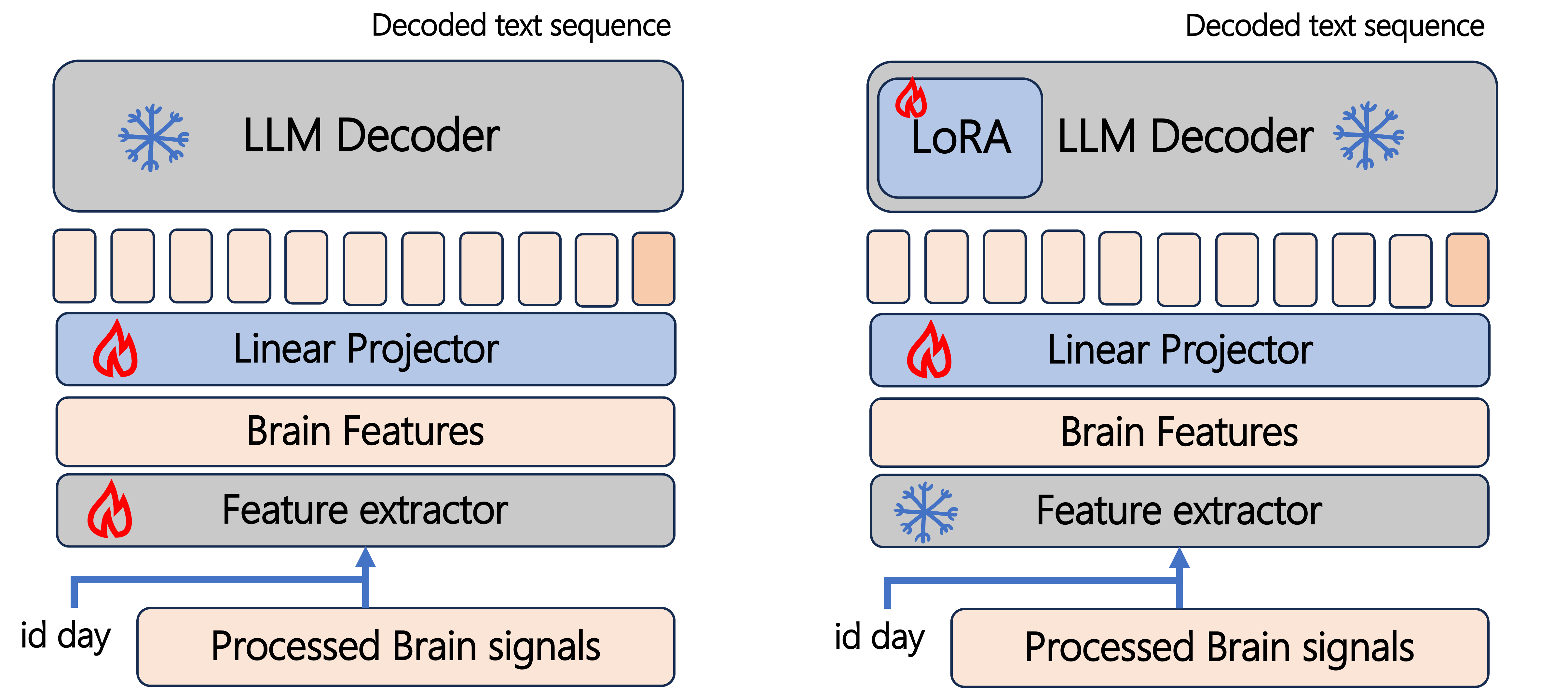}
    \caption{Multi-stage training strategy. In the modality alignment stage (left), the LLM is frozen. In the LLM finetuning stage (right), the feature extractor and most parameters of the LLM are frozen.}
    \label{fig:multistage}
\end{figure}

Our E2E framework employs a multi-stage training strategy, as illustrated in Figure \ref{fig:multistage}. The first stage involves modality alignment, which modifies the pre-trained feature extractor to map the brain signals into the space of specific Language Model (LLM) token embeddings. In this stage, we freeze the parameters of the LLM and only update the parameters of the feature extractor and the linear transformation layer. Following this, we conduct the LLM fine-tuning stage. Here, the goal is to fine-tune the LLM decoder to better understand the intermediate representations of the brain signals. During this stage, we freeze the parameters of the feature extractor and only update the parameters of the linear layer and the LLM decoder.

Our framework is designed in an E2E structure, directly modeling the relationship between raw brain recordings and text transcriptions. It takes brain signals and sequentially decodes sentences. We utilize negative log-likelihood loss as our objective function:
\begin{equation}
    \mathcal{L} (f,Y) = - \frac{1}{L}\sum_{i = 1}^{L} \log p_{f}(y_i|y_j,j<i)
\end{equation}

\section{Experiment}
\subsection{Dataset}
\begin{table}[h]
    \centering
    \resizebox{\linewidth}{!}{
    \begin{tabular}{*{6}{c}}
          \toprule
          & \multicolumn{2}{c}{number of sentences} & \multicolumn{2}{c}{number of unique words} & \multirow{2}{*}{word overlap}  \\
          \cmidrule(lr){2-3} \cmidrule(lr){4-5}\morecmidrules\
          & train & test &  train & test &  \\
          \midrule
          vocal & 6820	&720 &5866	&1361
	&0.796  \\
          silent & 1960 &160 &2152 &441 &0.742  \\
          \midrule
          all & 8780 & 880 &6413	&1519 & 0.810  \\
          \bottomrule
    \end{tabular}
    }
    \caption{The statistics of dataset. Word overlap refers to the percentage of words for test that appear in the training set.}
    \label{tab:dataset}
\end{table}

\vspace{-0.6cm}
We use the dataset released by Willet et al. \cite{willett2023high}, which contains the invasive brain recordings and text transcriptions collected from one participant on different days. The participant has bulbar-onset amyotrophic lateral sclerosis (ALS), retains limited ability to vocalize, and is unable to produce distinguishable speech by human ears. The neural activity is recorded by four microelectrode arrays as the participant is attempting to speak and is further converted to binned threshold crossings and spike band power of a total of 512 channels, among which 256 channels are used in our experiment. We present the statistics of the dataset in Table \ref{tab:dataset}. The vocal part was collected when the participant attempted to speak loudly, even though the speech was unintelligible. The silent part was collected when the participant spoke in her mind without making a sound.

\subsection{Data preprocessing}

We implement the same data preprocess by normalizing the brain signals channel-wise within each block. At the same time, we add white noise composed of element-wise and channel-wise Gaussian noise any time we access the brain signal during the training process before we perform Gaussian smoothing on each channel to prevent overfitting. During inference, Gaussian smoothing is used directly without adding white noise. It should be noted that we preserve the original form of the sentences during training and remove all punctuations and abbreviations during evaluation, which is closer to the real-world usage of brain-to-text decoding.

\subsection{Experimental settings}

For data preprocessing, the standard deviation of Gaussian noise is set to $1.0$ element-wise and $0.2$ channel-wise. During the training process, We use the AdamW optimizer \cite{loshchilov2018decoupled} and a linear learning rate decay scheduler with $400$ warm-up steps. The feature extractor is a 5-layer GRU with the hidden dimension $1024$. We set the initial learning rate to $0.02$ and add an L2 penalty with the coefficient $1e-5$ to pre-train the feature extractor with batch size $64$ for $400$ epochs. In the modality alignment stage, we set the initial learning rate $1e-3$ and train the framework with batch size $8$ for $100$ epochs. In the LLM finetuning stage, we set the initial learning rate among \{$5e-5$,$1e-5$\} for the LLM decoder and $1e-3$ for the linear layer, ensuring the linear layer adapts faster. We update all the parameters for small LLM (GPT-2) while we use the low-rank adaptation (LoRA) \cite{hu2021lora} with rank $8$ to reduce GPU memory usage and accelerate the training process for LLM with more parameters. The LLM finetuning stage lasts for $200$ epochs. 

We use word error rate (WER) as our evaluation metric, which necessitates the insertion, deletion, and substitution errors between the decoding hypothesis and the text annotation.

\section{Result and Discussion}

\begin{figure*}[pth]
    \centering
    \begin{minipage}[t]{0.48\textwidth}
        \centering
        \includegraphics[width=8cm]{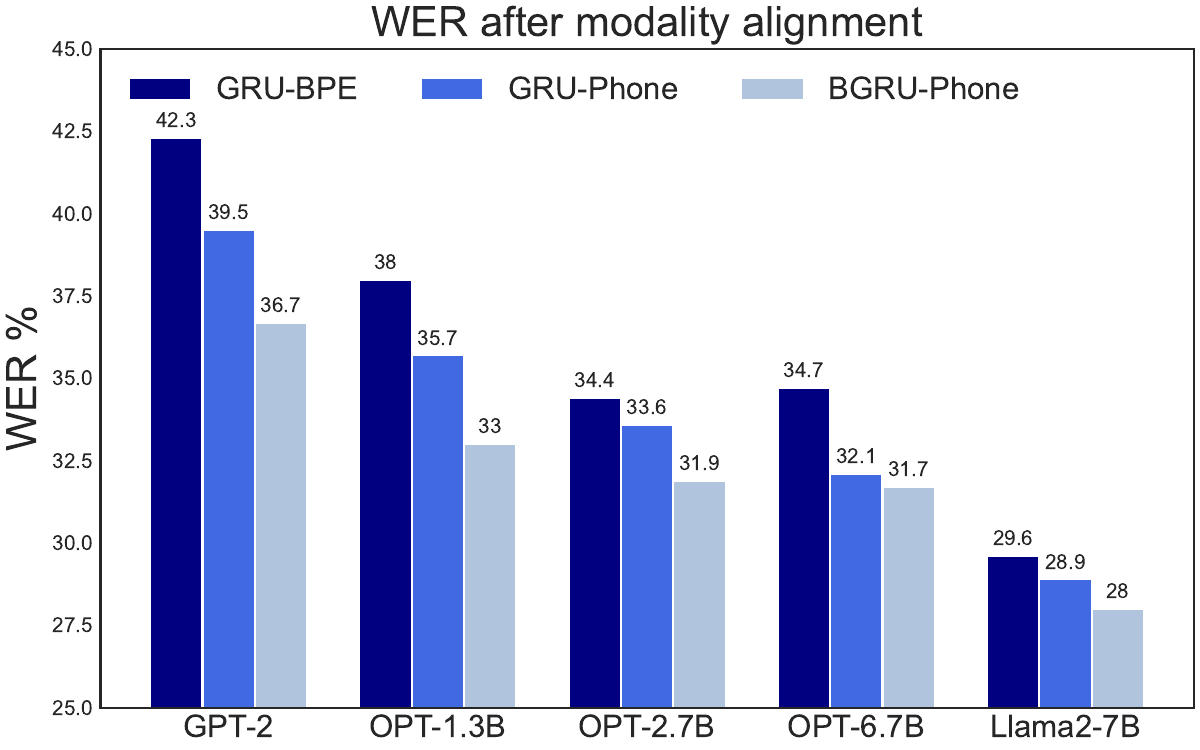}
    \end{minipage}
    \begin{minipage}[t]{0.48\textwidth}
        \centering
        \includegraphics[width=8cm]{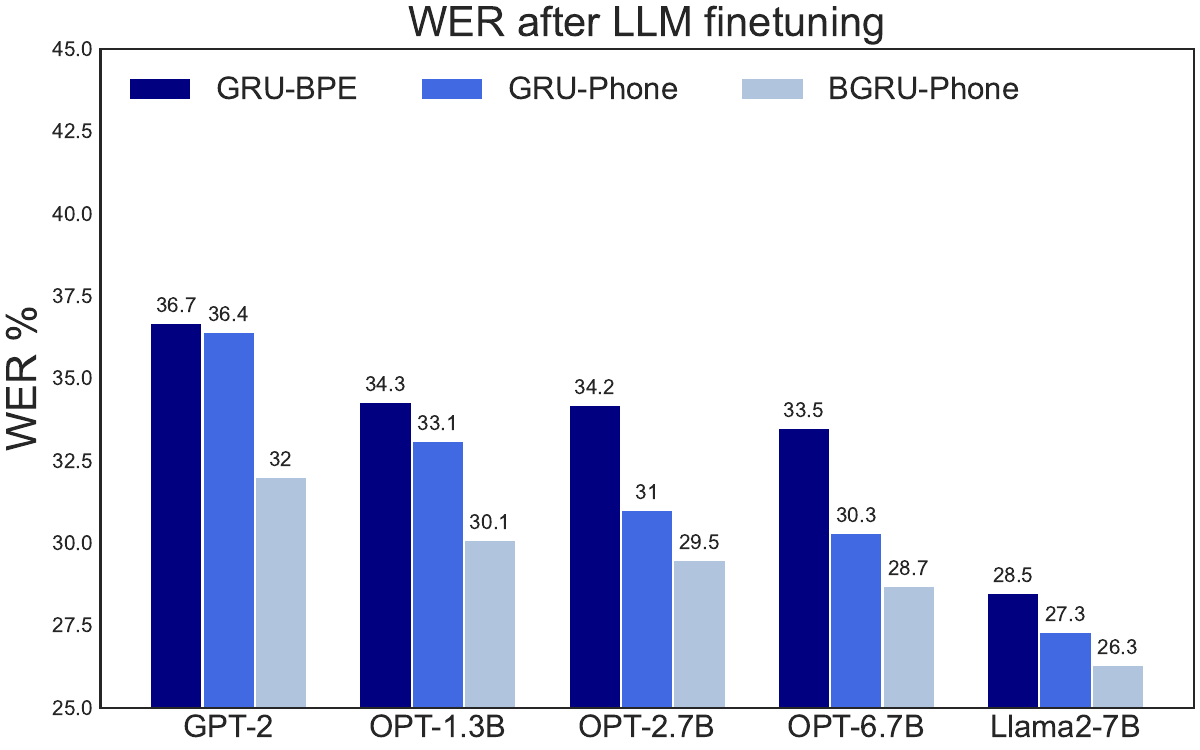}
    \end{minipage}
    \caption{WER on test dataset for different feature extractors and LLMs after modality alignment stage (left) and LLM finetuning stage (right). "BPE" and "Phone" refer to the pre-training task of the feature extractor: brain to BPE unit task and brain to phoneme task. "BGRU" means using bidirectional GRU as the structure of the feature extractor.}
    \label{fig:result}
\end{figure*}

\begin{table}[ht!]
    \centering
    \resizebox{\linewidth}{!}{
    \begin{tabular}{*{5}{c}}
          \toprule
          \multirow{2}{*}{\textbf{Feature Extractor}} & \multirow{2}{*}{\textbf{LLM Decoder}}& \multicolumn{3}{c}{\textbf{WER(\%)$\downarrow$}} \\
          \cmidrule(lr){3-5} \morecmidrules\
          & &  vocal & silent & all  \\
          \midrule
          \multirow{5}{*}{GRU-BPE}&GPT-2	&	36.8 &37.1 &36.9 \\
          &OPT-1.3B &34.1 &35.2&34.3\\
          &OPT-2.7B	&34.0 &35.2 &34.2\\
          &OPT-6.7B  &33.4 &34.1 & 33.5\\
          &Llama2-7B  &28.9 &26.9 & 28.5\\
          \midrule
          \multirow{5}{*}{GRU-Phone}&GPT-2	&36.4 &36.5 &36.4\\
          &OPT-1.3B	&33.2 &32.9 &33.1\\
          &OPT-2.7B  & 31.0 &30.8 &31.0\\
          &OPT-6.7B	&30.4 &29.6 &30.3\\
          &Llama2-7B &27.4 &26.7 &27.3\\
          \midrule
          \multirow{5}{*}{BGRU-Phone}&GPT-2	&32.2 &31.1 &32.0\\
          &OPT-1.3B  &30.1   &30.3 &30.1\\
          &OPT-2.7B  &29.4 &30.2 &29.5\\
          &OPT-6.7B &28.8 &28.2 &28.7\\
          &Llama2-7B  &\textbf{26.2} &\textbf{26.9} &\textbf{26.3}\\
          \midrule
          Hybrid \cite{willett2023high} & - & \textbf{23.8} & \textbf{24.7} & - \\
          \bottomrule
    \end{tabular}
    }
    \caption{Performance on vocal and silent sets separately.}
    \label{tab:result}
\end{table}

\vspace{-0.6cm}

We show our experimental results in Figure \ref{fig:result}, where we present the WER obtained on the test dataset after each stage of our training process using different feature extractors and LLM decoders. Although not mentioned in the figure, in the situation without LLM decoders, GRU-BPE can only achieve a WER of over 40\%, which is worse than any of our frameworks. This shows that textual information prior provided by LLMs is critical for the success of accurate decoding.

By comparing the final results obtained by using GRU pre-trained on different tasks, we find that the framework using GRU-Phone as the feature extractor is systematically better than the framework using GRU-BPE in both two stages, proving that the selection of the sub-word decomposition for the feature extractor pre-training tasks has an impact on the final performance of the framework. This suggests that the performance of the framework may be further ameliorated by using better sub-word decomposition. We also observe that the results using bidirectional GRU are better than those using normal GRU, implicating that the structure of the feature extractor is critical, and improvements may be made by further exploring the optimal structure to better capture the textual information from brain signals.

The modality alignment stage results show that by only adjusting the feature extractor and linear projection layer parameters, without updating the LLM decoder, a Word Error Rate (WER) close to 30\% can be achieved. For Llama2-7B, the WER achieved is lower than any other LLM candidates for every given feature extractor, demonstrating its superior capacity in handling input with different modalities. From the figure, we observe a clear tendency for the result of modality alignment to be better with a larger number of parameters, indicating that LLM with a larger scale understands brain signals better, even without any adaptation. The same trend is observed in the final results post-finetuning: as the number of LLM parameters increases, the WER decreases. The best result after finetuning is also achieved by using Llama2-7B.



The performance assessments of our models on both vocal and silent datasets following the fine-tuning phase with Large Language Models (LLMs) are presented in Table \ref{tab:result}. The most favorable outcome was achieved when employing BGRU-Phone for feature extraction alongside Llama2-7B as the LLM decoder. This configuration resulted in a Word Error Rate (WER) of 26.3\% on the test dataset, with specific WERs of 26.2\% for vocal datasets and 26.9\% for silent datasets. These figures are on par with those achieved by the hybrid model referenced in previous research, which reported WERs of 23.8\% and 24.7\% for vocal and silent datasets, respectively. Notably, in our optimal model setup, vocal datasets yielded a lower WER compared to silent datasets, aligning with findings from the cascade approach documented in \cite{willett2023high}. However, among all 15 evaluated models, only 8 demonstrated this trend of lower WERs for vocal datasets, suggesting that the observed difference might occur by chance. Further investigations are warranted to delve into the disparities between vocal and silent brain signals.

\section{Conclusion}

We propose the first E2E framework for large-vocabulary invasive brain signal decoding and achieve comparable results against the classical hybrid model, demonstrating the feasibility of this approach and providing a benchmark for E2E speech neuroprosthesis. We believe the switch from the cascade approach to the E2E paradigm is inevitable with a growing number of brain activity recordings with pronunciation. Our work can give insights to future researchers and contribute to the rapid development of speech-functional BCIs.



\newpage

\bibliographystyle{IEEEtran}
\bibliography{template}

\end{document}